\documentclass[journal]{IEEEtran}

\usepackage{cite}

\ifCLASSINFOpdf
  \usepackage[pdftex]{graphicx}
\else
\fi

\ifCLASSOPTIONcompsoc
 \usepackage[caption=false,font=normalsize,labelfont=sf,textfont=sf]{subfig}
\else
 \usepackage[caption=false,font=footnotesize]{subfig}
\fi

\usepackage{multirow}
\usepackage{xcolor}

\begin{document}

% \title{Self-reliant Auto-Encoder Neural Network for JPEG restoration}

% \title{Deep Residual Auto-Encoder for JPEG compression artifact reduction}

% \title{Deep Residual Auto-Encoder for the Restoration of JPEG-Compressed Images}

% \title{Deep Residual Auto-Encoder for JPEG restoration}

% \title{QF-resilient Deep Residual Auto-Encoder for JPEG restoration}

% \title{Deep Residual Auto-Encoder for JPEG restoration: a compression quality factor independent model}

% \title{Deep Residual Auto-Encoder for JPEG restoration: a single, quality factor-resilient model}

% \title{Implicit compression-estimation Deep Residual Auto-Encoder for JPEG restoration}

% \title{Deep Residual Auto-Encoder for JPEG restoration: a compression quality factor independent method}
 
% \title{Deep Residual Auto-Encoder for JPEG restoration: a compression-rate independent method}

\title{Deep Residual Autoencoder for quality independent JPEG restoration}

% handle multiple compression qualities in a single model (da "Compression Artifacts Removal Using Convolutional Neural Networks")

%We have shown that a network trained for a specific quality factor is resilient to the QF used compress the input image—a single network trained for QF 60 provides a PSNR gain of more than 1.5 dB over the wide QF range from 40 to 76 (da CAS-CNN: A Deep Convolutional Neural Network for Image Compression Artifact Suppression)


% \author{Simone~Zini,~\IEEEmembership{XXX,~XXX,}
%         Simone~Bianco,~\IEEEmembership{XXX,~XXX,}
%         and~Raimondo~Schettini,~\IEEEmembership{XXX,~XXX}% <-this % stops a space
% \thanks{M. Shell was with the Department
% of Electrical and Computer Engineering, Georgia Institute of Technology, Atlanta,
% GA, 30332 USA e-mail: (see http://www.michaelshell.org/contact.html).}% <-this % stops a space
% \thanks{J. Doe and J. Doe are with Anonymous University.}% <-this % stops a space
% \thanks{Manuscript received April 19, 2005; revised August 26, 2015.}}

\author{Simone~Zini,
        Simone~Bianco
        and~Raimondo~Schettini% <-this % stops a space
\thanks{S. Zini, S. Bianco, and R. Schettini are with the Department of Informatics, Systems and Communication, University of Milano-Bicocca, 20126 Milan, Italy (email: s.zini1@campus.unimib.it; simone.bianco@unimib.it; schettini@disco.unimib.it).}% <-this % stops a space
% \thanks{J. Doe and J. Doe are with Anonymous University.}% <-this % stops a space
% \thanks{Manuscript received April 19, 2005; revised August 26, 2015.}
}

\maketitle

\begin{abstract}

In this paper we propose a deep residual autoencoder exploiting Residual-in-Residual Dense Blocks (RRDB) to remove artifacts in JPEG compressed images that is independent from the Quality Factor (QF) used. 
The proposed approach leverages both the learning capacity of deep residual networks and prior knowledge of the JPEG compression pipeline. The proposed model operates in the YCbCr color space and performs JPEG artifact restoration in two phases using two different autoencoders: the first one restores the luma channel exploiting 2D convolutions; the second one, using the restored luma channel as a guide, restores the chroma channels explotining 3D convolutions.

Extensive experimental results on three widely used benchmark datasets (i.e. \textsc{LIVE1}, \textsc{BDS500}, and \textsc{CLASSIC-5}) show that our model is able to outperform the state of the art with respect to all the evaluation metrics considered (i.e. PSNR, PSNR-B, and SSIM). This results is remarkable since the approaches in the state of the art use a different set of weights for each compression quality, while the proposed model uses the same weights for all of them, making it applicable to images in the wild where the QF used for compression is unkwnown. Furthermore, the proposed model shows a greater robustness than state-of-the-art methods when applied to compression qualities not seen during training.

%abstract abstract abstract abstract abstract abstract abstract abstract abstract abstract abstract abstract abstract abstract abstract

\end{abstract}
%In this paper, we design a Deep Dual-Domain (D3) based fast restoration model to remove artifacts of JPEG compressed images. It leverages the large learning capacity of deep networks, as well as the problem-specific expertise that was hardly incorporated in the past design of deep architectures. For the latter, we take into consideration both the prior knowledge of the JPEG compression scheme, and the successful practice of the sparsity-based dual-domain approach. We further design the One-Step Sparse Inference (1-SI) module, as an efficient and light-weighted feed-forward approximation of sparse coding. Extensive experiments verify the superiority of the proposed D3 model over several state-of-the-art methods. Specifically, our best model is capable of outperforming the latest deep model for around 1 dB in PSNR, and is 30 times faster.


% \fi
\IEEEpeerreviewmaketitle

\section{Introduction}
% for IEEE journal papers produced under \LaTeX\ using
% IEEEtran.cls version 1.8b and later.

% \begin{figure}[!t]
% \centering
% \includegraphics[width=3in]{comp.pdf}
% \caption{Comparison.}
% \label{fig_comp}
% \end{figure}

\begin{figure}[!t]
\centering
\includegraphics[width=\linewidth]{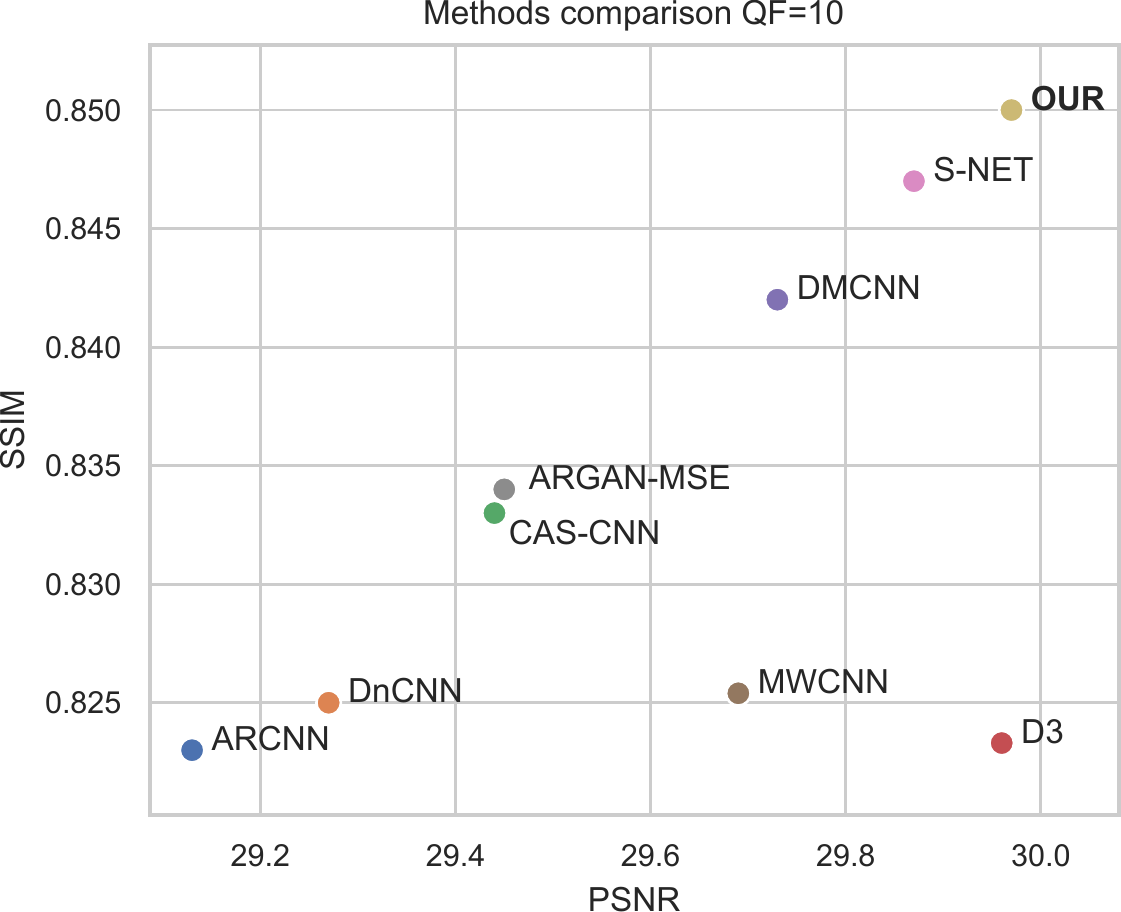}
\caption{PSNR-SSIM comparison of the state-of-the-art-models and our proposed method. For both metrics higher value means better visual results.}
\label{fig_qf_10}
\end{figure}

Image compression represents a very active research topic due to the high impact of the data in a big amount of  fields, from image sharing on the web to the most specific applications involving the acquisition of images and transfer to elaboration nodes.

Specifically, image compression refers to the task of representing images using the smallest storage space possible. 

Compression algorithms play a key role for saving space and bandwidth for the memorization and transfer of large amount of images.
Two different compression paradigm exist: the former is 
lossless image compression, where the compression rate is limited
by the requirement that the original image must be perfectly recovered; the latter, more diffused, is lossy image compression, where higher compression rates are possible at the cost of some distortion in the recovered image.
Among the lossy compression algorithms, the most diffused and used is the JPEG compression algorithm.

The JPEG compression algorithm first converts the original RGB image into YCbCr color space and processes the luma and chroma channels separately. It divides the luma channel of an input image into non-overlapping $8\times8$ blocks and performs the Discrete Cosine Transform (DCT) on each block separately, while downsampling the chroma components with a bilinear filter. The DCT coefficients obtained from the luma channel are then quantized based on \textit{quantization tables} and adjusted using the user-selected \textit{quality factor}. The image is then reconstructed from the quantized DCT coefficients by using the inverse DCT. The described JPEG encoding operation introduces three kinds of artifacts in the recovered images, related to the quality factor used for the compression:
% \begin{itemize}
%     \item[-] Blocketization artifacts;
%     \item[-] Ringing artifacts;
%     \item[-] Blurred low-frequencies areas.
% \end{itemize}
% The first kind of artifacts comes from the recombination of the $8\times8$ blocks, that are compressed without considering the adjacent blocks, ringing effects along the edges are related to the coarse quantization of the high-frequencies components and the blurred effect on the images is also related to the compression of the high-frequencies in the DCT domain.
i) blocketization artifacts, which come from the recombination of the $8\times8$ blocks, that are independently compressed without considering the adjacent blocks;
ii) ringing artifacts, which are most visible along the edges and are related to the coarse quantization of the high-frequencies components;
iii) blurred low-frequencies areas, which is also related to the compression of the high-frequencies in the DCT domain.

The presence of these kinds of artifacts represents a problem since the general quality of the images is degraded resulting unpleasing for normal users for generic applications (e.g. projection, print, etc.), or even useless for computer vision applications where the loss of information can be potentially critic for the task \cite{dodge2016understanding,bianco2016robust}.

With the purpose of reducing these artifacts, in the last years a lot of JPEG artifact reduction algorithms have been proposed. These methods include both traditional image processing pipelines~\cite{list2003adaptive,reeve1984reduction,wang2013adaptive,ahmed1974discrete,foi2006pointwise,jancsary2012loss} and machine learning approaches~\cite{dong2015compression,zhang2017beyond,cavigelli2017cas,wang2016d3,galteri2017deep,Liu_2018_CVPR_Workshops,zhang2018dmcnn,zheng2018s}, both making great steps in the restoration of corrupted images. However, these methods suffer from two main limits: the first one is that they need to train a different model for each possible quality factor (QF), making them not generally applicable to general images downloaded from the web unless the QF used for compression is known; the second one, is that the great majority of methods in the state of the art restores just the luma channel or do not fully exploit the knowledge about the JPEG compression pipeline. 
%, \textcolor{red}{linked to the ability of correctly reconstruction information in RGB color space while dealing with different compression quality factors}.

To address these problem we propose a new method for the restoration of JPEG compressed images in YCbCr color space, based on machine learning, specifically on convolutional autoencoders. 
The proposed approach consists in two deep autoencoders respectively used for luma and chroma restoration, %capable to restore structures and colors in JPEG degraded images, 
that are able to restore images independently from the quality factor used for the compression. 
The main contribution are the following:
\begin{itemize}
    \item[-] the design of a method for the restoration of JPEG compression artifact that is independent from the QF used;
    \item[-] the design of a model trainable end-to-end that fully exploits knowledge about JPEG compression pipeline;
    \item[-] a thorough comparison with the state of the art on three standard datasets at fixed QFs;
    \item[-] an analysis of robustness of restoration results at QFs not used for training.
\end{itemize}

%The rest of the paper is organized as follows
\section{Related~Works}
% The task of JPEG compression artifacts removal has been attempted from different points of view in the past years. 
The task of JPEG compression artifacts removal has been faced in different ways in the past years. 
The existing proposed methods can be broadly classified into two groups: traditional image processing methods and learning based methods.

%%% Processing based
% Within the first group have been proposed both spatial and frequency domain based methods.
To the first group belong methods based on traditional image processing techniques working both in the spatial and in the frequency domain.
For spatial domain processing different kinds of filters have been proposed, with the intent of restoring specific areas of the images such as edges \cite{list2003adaptive}, textures \cite{reeve1984reduction}, smooth regions\cite{wang2013adaptive}, etc.
Algorithms usually rely on information obtained by the application of the Discrete Cosine Transform (DCT) transform \cite{ahmed1974discrete}. SA-DCT, proposed by Foi \textit{et} al. \cite{foi2006pointwise}, attempts to reconstruct an estimate of the signal using the DCT of the original image together with the spatial information contained in the image itself. However SA-DCT is not capable to reproduce details like sharp edges or complex textures. To overcome this limit different restoration oriented methods have been proposed, like the Regression Tree Fields based method (RTF)~\cite{jancsary2012loss}. The RTF uses the results of SA-DCT to restore images, taking advantage of a regression tree field model.

%%% Machine Learning based
Following the success of the application of Deep Convolutional Neural Networks (Deep-CNNs) in image processing tasks, such as image denoising \cite{zhang2017beyond} and Single-Image Super-Resolution \cite{dong2016image}, Deep-CNNs have been applied with success to JPEG compression artifact removal task. The basic idea behind Deep-CNNs is to learn a function to map a set of images from an input distribution, to the desired output one.
In the artifact removal case the objective is to map degraded images into a distribution without the presence of the noise. The trained neural network obtained at the end of the training process represent an approximation of the desired function for the translation of the images from a distribution to another one.

The first attempt with this kind of models has been done by Dong \textit{et al.}\cite{dong2015compression} who proposed the ARCNN, a model inspired by SRCNN \cite{dong2016image}, a neural network for Super-Resolution. This first attempt has been followed by DnCNN \cite{zhang2017beyond}, a CNN for general denoising task that has also been used on JPEG compressed images, and CAS-CNN \cite{cavigelli2017cas}, a model proposed by Cavigelli \textit{et al.}, who presented a much deeper model capable to obtain higher quality images.
Wang \textit{et al.} proposed D3 \cite{wang2016d3}, a deep neural network that adopts JPEG-related priors to improve reconstruction quality which obtained an improvement in speed and performances with respect with to the previous models.
In 2017, Galtieri \textit{et al.}\cite{galteri2017deep} developed a generative adversarial network (GAN)\cite{NIPS2014_5423} for artifact removal and texture reconstruction.

In 2018 a bunch of new models for JPEG artifact removal has been presented, showing interesting improvements in the results quality. Liu \textit{et al.} \cite{Liu_2018_CVPR_Workshops} proposed a Multi-level Wavelet CNN (MWCNN), a model based on the U-Net architecture \cite{ronneberger2015u}, trained and used for multiple tasks: compression artifact removal, denoising and super-resolution. Zhang \textit{et al.} \cite{zhang2018dmcnn} developed DMCNN, a Dual-Domain Multi-Scale CNN, which gains higher results quality than the previous works, by using both pixel and frequency (\textit{i.e.} DCT) domain information. Lastly S-Net, the most recent method by Zheng \textit{et al.} \cite{zheng2018s} proposed a ``greedy loss architecture" to train deeper models capable to outperform the previous state-of-the-art.

% However, the methods proposed until now suffers from two limits:
% each machine learning model needs to know the JPEG compression quality factor of each input image to properly restore a compressed image and also are capable to restore only the luminance channel, without considering the chroma components (except for S-Net). 

\section{Proposed~Method}
%\textcolor{red}{\bf{TBD:} luminance vs luma, chrominance vs chroma}

The method in the state of the art mainly suffer from two limits:
the first one is that each machine learning model needs to know the JPEG compression Quality Factor (QF) of each input image to properly restore a compressed image; the second one is that the great majority of them are capable to restore only the luma channel without considering the chroma components, and the only one that recovers all three channels \cite{zheng2018s} does not fully exploit theoretical knowledge of the JPEG compression pipeline. 
% and the only one recovering both luma and chr (except for S-Net). 

% Problemi principali: qualitá specifiche, elaborazione del solo canale Y
% Our main purpose is to overcome two problems, presented at the end of the previous section, that are not considered by the previous works:
% \begin{itemize}
%     \item prior knowledge of JPEG compression quality factor ($QF$) of the input images;
%     \item luma channel specific restoration.
% \end{itemize}

In this work we propose a method able to overcame both these problems.
The first problem has to do with the way the models are trained: all of the previous existing methods make the implicit assumption that the compression quality factor QF used to compress the input images is known. In fact, most of the previous models present networks trained on datasets compressed on specific quality factors (the most common being $\rm{QF}=10,20,30$ and $40$).
This way of training the models leads to two limits:
\begin{itemize}
    \item[-] the models are capable to correctly restore only images at a specific QF, with the consequence that a specific training for each quality factor is needed;
    \item[-] the QF used for the compression of the images is needed in order to train a model and correctly restore the images: this is usually a not known information for images coming from unknown sources (e.g. downloaded from the web), thus largely limiting the usability of the model.
\end{itemize}

In order to overcome the necessity to know the compression quality factor, we train our model on a dataset containing images compressed at different QFs: this will make the model more generic and able to restore images taken in the wild, i.e. without knowing the actual QF used. This objective poses a challenge, since the training of such a quality independent model is much harder than training on a single quality factor. 

The second problem concerns the way the previous models restore the images: all of the previous state-of-the-art methods are trained on the luma channel (\textit{Y} channel of the \textit{YCbCr} space) of the images. This approach is based on the fact that the JPEG compression algorithm applies the DCT to the \textit{Y} channel, introducing ringing and blocketization artifacts on the luma channel, while the other \textit{Cb} and \textit{Cr} channels are just sub-sampled the bicubic interpolation.
The design and training of a model for the specific restoration of the luma component and its subsequent application for the restoration of the chroma components (as done for example by ARCNN \cite{dong2015compression}), introduces chromatic aberrations and artifacts in the final result.
S-Net\cite{zheng2018s} is the only method considering this problem and instead of training a model for the restoration of just the luma component, it takes as input a full RGB image and recovers a full RGB images as output.
% it works on 3-channels RGB images instead of working separately on luma, attempting to restore both structures and chroma in RGB space.

To overcome this second limit and obtain better results we exploit the knowledge of how the JPEG compression pipeline works and propose the use of two models for the image restoration in \textit{YCbCr} space: the first model restores the \textit{Y} channel; the second model then uses the result as a \textit{Structure Map} (i.e. a guide) for the restoration of the chroma components. 
A schematic representation of the proposed method is depicted in Figure \ref{fig_ov}.

% \textcolor{black!20}{We present a Deep Autoencoder CNN, based on a new revisited version of the Residual Blocks from ResNet\cite{he2016deep}, which has been used with some differences for both luma and chroma image restoration.}

%%%%%%%%%%%%%%% CB e CR
\subsection{Luma and chroma Restoration Model}
The vast majority of learning based methods for JPEG compression artifact removal in the state of the art  \cite{dong2015compression, zhang2017beyond, cavigelli2017cas, wang2016d3, Liu_2018_CVPR_Workshops, zhang2018dmcnn} focus exclusively on the luma component of the images. 
%compressed by the JPEG compression algorithm with the use of DCT over the original images. 
Generally these methods perform the compression artifact removal 
%remove the The general way of operating for the blocking and ringing artifact removal is to work 
working on the \textit{Y} channel of the images, after converting them in \textit{YCbCr} color space. The learned model in some cases is then applied as is also on \textit{Cb} and \textit{Cr} channels (e.g. \cite{dong2015compression}).
These approaches do not take in consideration the chroma aspects of the images, generating results with aberrations in RGB space and low perceptual quality.

\begin{figure*}[!t]
\centering
\includegraphics[width=\linewidth]{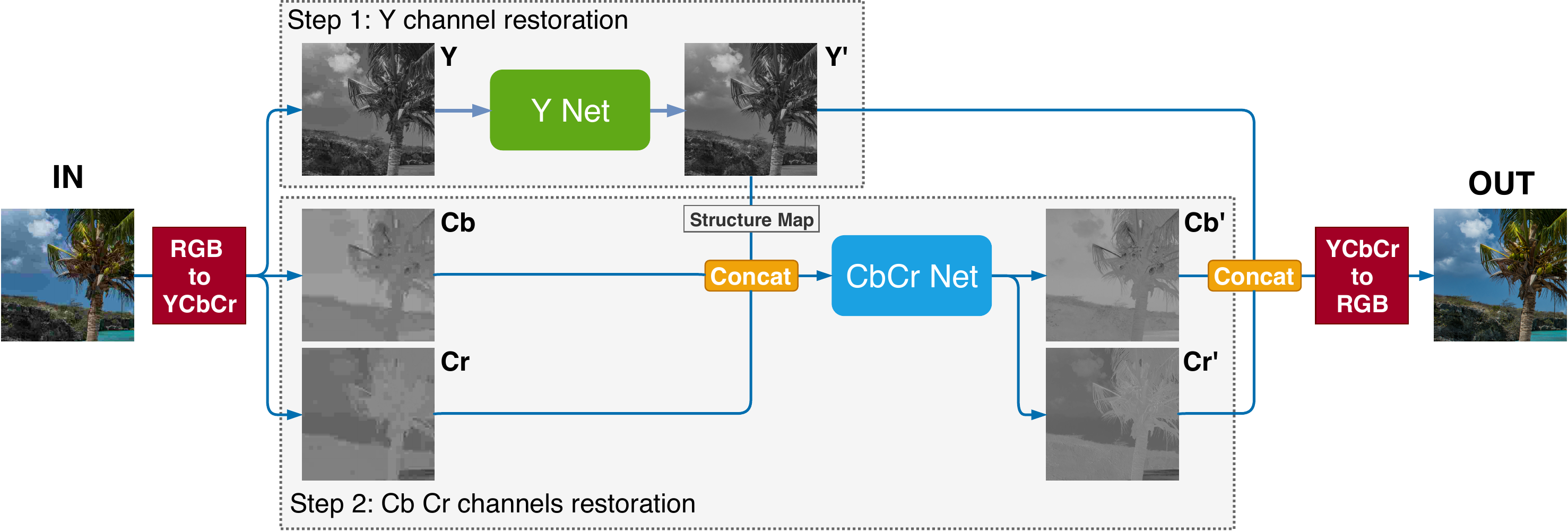}
\caption{Schematic representation of the proposed method: the input image is first converted to \textit{YCbCr} color space. The \textit{Y} channel is restored with the Y-net and the result \textit{Y'} is concatenated with the original \textit{CbCr} channels to restore \textit{Cb'Cr'} with the CbCr-net. Restored \textit{Y'Cb'Cr'} channels are then converted back to RGB color space.}
\label{fig_ov}
\end{figure*}

% \textcolor{red}{
% Keeping this in mind we propose a system for restoring both luma and chroma components of the compressed images (see Figure \ref{fig_ov}). The method consists of two steps: first we use the artifact removal model, called \textit{Y Net}, on the \textit{Y} channel of the input image; then the \textit{Cb} and \textit{Cr} channels of the original image are restored by a secondary model, another autoencoder with some differences in the structure called \textit{CbCr Net}, using the previously predicted \textit{Y}$'$ channel as a map of the structures present in the image.}

\begin{figure}[!t]
\centering
%\hfill
\subfloat[Original]{\includegraphics[width=0.46\linewidth]{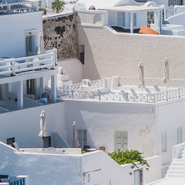}%
\label{fig_original}}\hfill
\subfloat[Compressed ($\rm{QF}=10$)]{\includegraphics[width=0.46\linewidth]{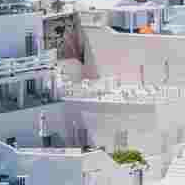}%
\label{fig_input}}
\vspace{-0.2truecm}
%\hfill
\subfloat[Original \textit{Cb}]{\includegraphics[width=0.46\linewidth]{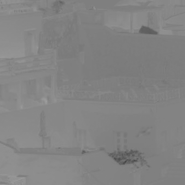}%
\label{fig_originalcb}}\hfill
\subfloat[Compressed \textit{Cb} ($\rm{QF}=10$) ]{\includegraphics[width=0.46\linewidth]{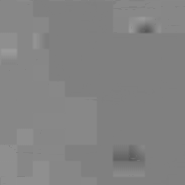}%
\label{fig_inputcb}}
\vspace{-0.2truecm}
%\hfill
\subfloat[Original \textit{Cr}]{\includegraphics[width=0.46\linewidth]{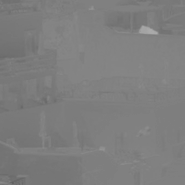}%
\label{fig_originalcr}}\hfill
\subfloat[Compressed \textit{Cr} ($\rm{QF}=10$)]{\includegraphics[width=0.46\linewidth]{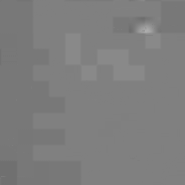}%
\label{fig_inputcr}}
\hfill
\caption{Visual example of how the JPEG compression algorithm, when operating
with very low compression quality factors changes the colors of the input images in two different ways: hue change and spatial location change.}
\label{fig_color_error}
\end{figure}

Moreover the JPEG compression algorithm, when operating with very low compression quality factors, such as $\rm{QF}<20$, tends to change the colors of the input images in two different ways: hue change and spatial location change. As can be seen in Figure \ref{fig_color_error}, in the compressed version of the \textit{Cb} and \textit{Cr} channels, as expected the color resolution is reduced, and also, for some elements, the color position does not correspond to the one in the original uncompressed image.

Keeping the above considerations in mind we propose a method for restoring both luma and chroma components of the compressed images (see Figure \ref{fig_ov}). The method consists of two steps: the first step, after the conversion of the input image into \textit{YCbCr} color space, involves the restoration of the \textit{Y} channel alone, using a first model named LumiNet, and produces \textit{Y'} as output. 
The second step concatenates \textit{Y'CbCr} along the channel dimension and uses a second model named ChromaNet, to restore the \textit{CbCr} channels. This second step uses \textit{Y'} as a map of the structures present in the image (i.e. a sort of guide) to condition the second network to recover the color hue and contours, and produces \textit{Cb'Cr'} as output. 
The final output is obtained by concatenating \textit{Y'Cb'Cr'} and converting them back to RGB.
Both LumiNet and ChromaNet are two different deep CNN Autoencoders both exploiting a new revisited version of the Residual Blocks \cite{he2016deep}.

%The JPEG compression algorithm, when operating with very low compression quality factors, such as $\rm{QF}<20$, tends to change the colors of the input images in two different ways: hue change and spatial location change. As can be seen in Figure \ref{fig_color_error}, in the compressed version of the \textit{Cb} and \textit{Cr} channels, as expected the color resolution is reduced, and also, for some elements, the color position does not correspond to the one in the original uncompressed image.
%In order to restore the color in the \textit{Cb Cr} channels taking care of these two points, we attempt to use the predicted \textit{Y}$'$ channel as a map to condition the second network to recover the color hue and contours, since in the luma channel is contained the information about the structures of the image.
%After restoring the \textit{Y} channel of the compressed image, the \textit{Cb} and \textit{Cr} original channels are concatenated, along the channel dimensions, with the \textit{Y}$'$ restored luma channel and feeded in the \textit{CbCr Net}.

%%%%%%%%%%%%%%% Architettura
\subsection{Deep Residual Autoencoder Architecture}

\begin{figure*}[!t]
\centering
\includegraphics[width=0.9\textwidth]{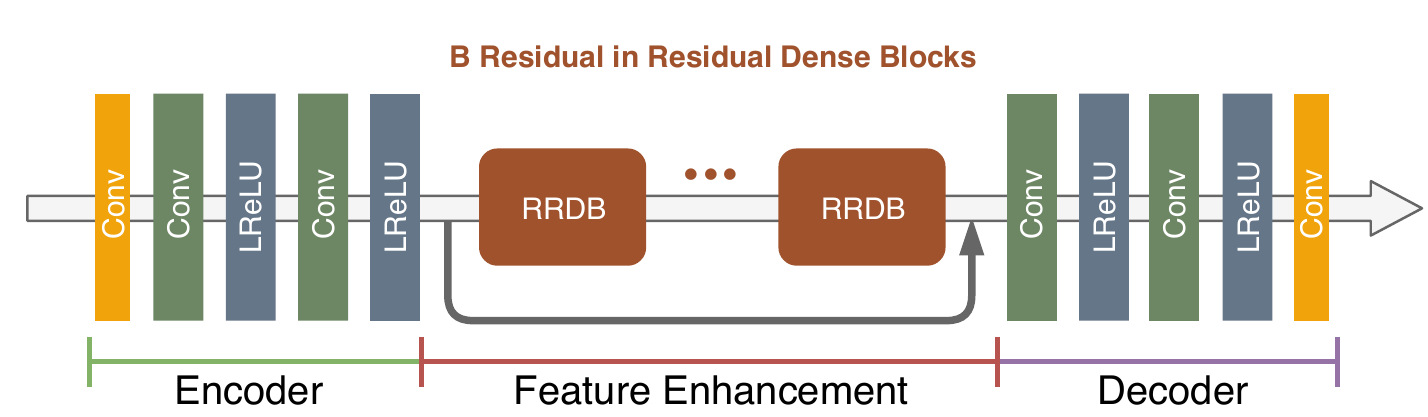}
\caption{Graphical representation of the architecture of the autoencoders used for both the luma and chroma restoration.}
\label{fig_arch}
\end{figure*}
%\textcolor{red}{\bf TBD: Auto-Encoder vs autoencoder}

Autoencoder architectures have been widely used in image processing tasks like image-to-image translation \cite{pix2pix2017}, Super-Resolution \cite{zeng2017coupled}, image inpainting\cite{xie2012image} and rain removal \cite{qian2018attentive}. Autoencoders generally present a structure made by three parts: the encoder, which extracts features from the $n$-dimensional input (usually 1 or 3 channels); a central part, that performs feature processing; and the final decoder, which decodes the processed features into the output image having the desired dimensions.
Figure \ref{fig_arch} shows a schematic representation of the proposed model, while a more detailed description of its architecture is reported in Table \ref{tab_arch}.

\begin{figure}[!t]
\centering
\includegraphics[width=\linewidth]{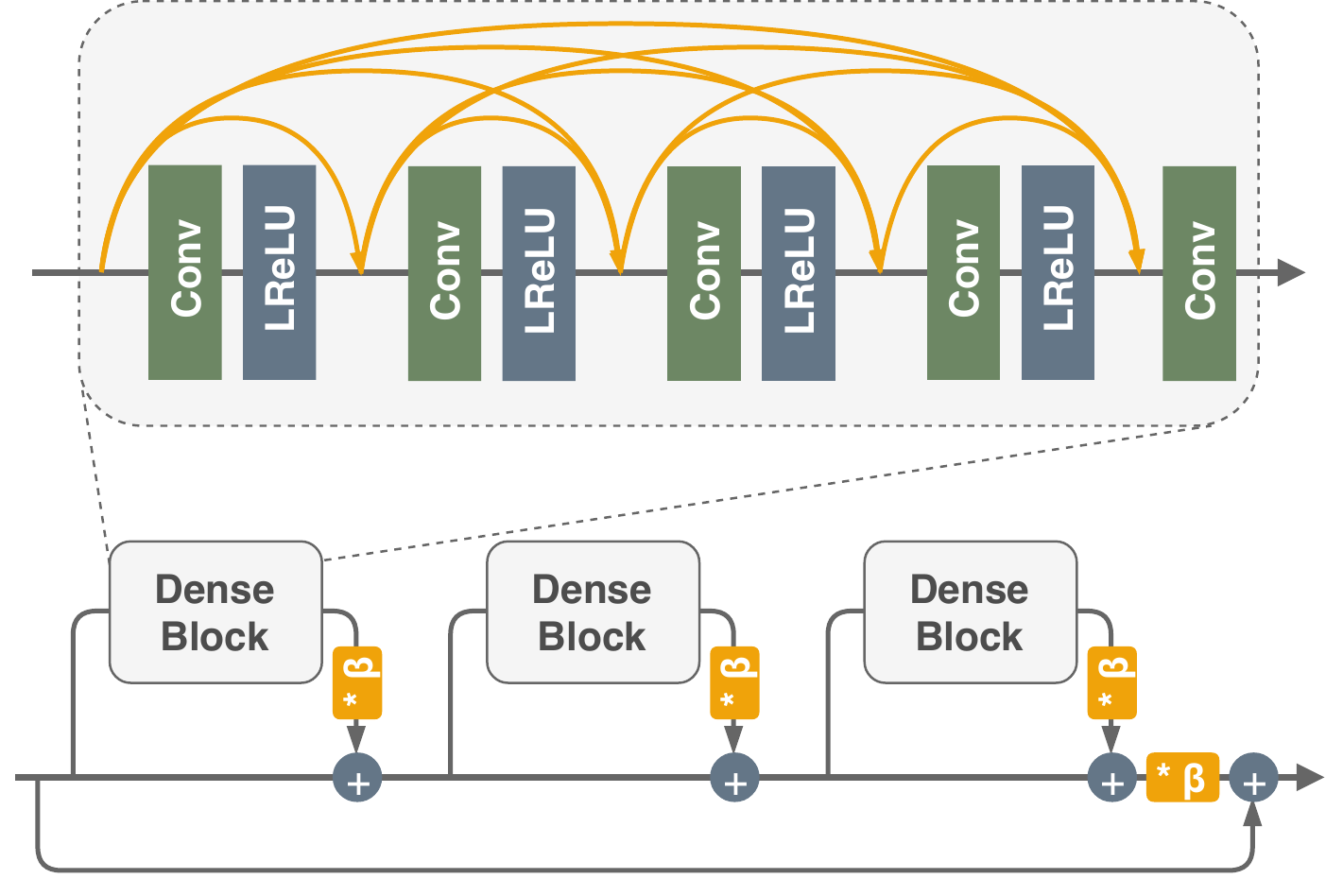}
\caption{Schematic representation of the architecture of the Residual-in-Residual Dense Block (RRDB) \cite{wang2018esrgan}.}
\label{fig_rrdb}
\end{figure}

The encoder, which consists of two convolutions followed by Leaky ReLU activations, is followed by a central part for feature enhancement consisting in a sequence of  \textit{Residual-in-Residual Dense Blocks} (RRDB) \cite{wang2018esrgan}, a modified version of the well known residual blocks originally introduced in the ResNet architecture\cite{he2016deep}, 
that have been shown to perform well in other image processing tasks, e.g. image super-resolution \cite{lim2017enhanced,wang2018esrgan}. 
The RRDBs blocks combine multi-level residual learning and dense connection architecture: the RRDBs are designed without the use of the Batch Normalization and the application of the residual learning on different levels. The RRDBs are shown in Figure \ref{fig_rrdb}: each RRDB is made of five \textit{Dense Blocks}, which use only convolutions with Leaky ReLUs activation and dense skip connection structures, combined together with other skip connections.
Finally, the decoder is designed in a symmetrical way with respect to the encoder part.

The same architecture has been used for both the networks for luma and chroma restoration, but with some differences:
\begin{itemize}
\item[-] different depth in terms of number of RRDBs used in the central part;
\item[-] different feature extraction from the input in the encoder part.
\end{itemize}

For the restoration of the luma (\textit{Y} channel) the number of central RRDBs is set to five, while for the \textit{CbCr} restoration the number of RRDB is decreased to three. The second and more important difference is in the first layer of the \textit{CbCr version} of the network, which is a \textit{3-dimensional convolutional layer}. Considering that the input of the \textit{CbCr-Net} is the concatenation (along the channel dimension) of the restored \textit{Y'} channel with the \textit{Cb} and \textit{Cr} channels, we decided to use a 3D convolution to make the model capable to correlate information about color and structures with the use of the same kernels for all the information coming from the three input channels.
The output of this second network are the two restored \textit{Cb} and \textit{Cr} channels, which are then concatenated with the restored \textit{Y'} channel, in order to obtain the complete restored image.

\begin{table}[!t]
\centering
\caption{Detailed architecture of the autoencoders used for both the luma and chroma restoration. The number of RRDBs is $B=5$ for the Y-Net and $B=3$ for the CbCr-Net.}
\begin{tabular}{|lccc|}
\hline
                         & Layer  & Filter size, Stride, Padding & output channels \\\hline\hline
                         & Conv2D & 3x3, 1, 1                    & 64              \\\hline
\multirow{4}{*}{Encoder} & Conv2D & 5x5, 1, 2                    & 128             \\
                         & LReLU  & -                            & 128             \\
                         & Conv2D & 3x3, 1, 1                    & 64              \\
                         & LReLU  & -                            & 64              \\\hline
\multicolumn{4}{|c|}{\multirow{2}{*}{RRDB x B}}                                    \\
\multicolumn{4}{|c|}{}                                                             \\\hline
\multirow{4}{*}{Decoder} & Conv2D & 3x3, 1, 1                    & 128             \\
                         & LReLU  & -                            & 128             \\
                         & Conv2D & 5x5, 1, 2                    & 64              \\
                         & LReLU  & -                            & 64              \\\hline
                         & Conv2D & 3x3, 1, 1                    & 1               \\\hline
                         & Tanh   & -                            & 1               \\\hline
\end{tabular}
\label{tab_arch}
\end{table}

%%%%%%%%%%%%%%% RRDB
%\textcolor{red}{Based on the last results achieved in Super-Resolution task, in particular by Lim \textit{et al.}\cite{lim2017enhanced} with EDSR net and successively by Wang \textit{et al.}\cite{wang2018esrgan} with ESRGAN, we decided to adopt a new version of the Residual Blocks, optimized for enhancement and generation of images.}

%\textcolor{red}{Following the work done by Wang \textit{et al.} for the design of the ESRGAN, winner at the \textsc{PIRM2018} challenge \cite{blau20182018}, we construct our model keeping some key points:}
In order to improve the quality of the generated results, as well as to make the training process more stable, the proposed architecture include the following design choices:
\begin{itemize}
    \item[-] removal of Batch Normalization (BN) layers from the Residual Blocks;
    \item[-] use of a \textit{residual scaling} parameter in each Residual Block;
    \item[-] initialization of the model weights using a scaled version of the Kaiming initialization\cite{he2015delving}.
\end{itemize}

The removal of the batch normalization layers has been proved, in image Super-Resolution \cite{lim2017enhanced} and image deblurring \cite{nah2017deep} tasks, to increase the performances for the generation of images in terms of quality indexes (PSNR and SSIM \cite{wang2004image}). 
%In EDSR model the BN layers have been simply removed from the traditional residual blocks, while in the ESRGAN the new RRDB were designed. 
%\textcolor{red}{These new blocks combine multi-level residual learning and dense connection architecture: the RRDBs are designed without the use of the Batch Normalization and the application of the residual learning on different levels. The RRDBs are shown in Figure \ref{fig_rrdb}: each RRDB is made of five \textit{Dense Blocks}, which use only convolutions with Leaky ReLUs activation and dense skip connection structures, combined together with other skip connections.}
The removal of the BN layers, which improve the stability of the training and the generated image appearance, makes on the other hand the training of deep networks more difficult. To solve that issues two solutions have been proved to work well: the so called residual scaling (in our model set to $0.2$), to scale each residual in order to not magnify the input image in a wrong way, and a small weight initialization, obtained by the application of the Kaiming initialization, presented by He \textit{et al.}\cite{he2015delving}, scaled by a factor $0.1$. As can be seen in Figure \ref{fig_rrdb} the residual scaling is applied on the higher level of the residual learning architecture, i.e. 
on the output of each dense block and at the end of the RRDBs.
\section{Experimental~Setup}
The training of the proposed method leads to two different Deep-CNNs respectively for the restoration of the luminance and chroma components of JPEG compressed images at generic quality (i.e. QFs). In order to evaluate the results, our models have been compared with the state of the art in four different experimental setups:
\begin{enumerate}
    \item \textit{known QF} luminance restoration:  comparison with the state-of-the-art methods which work only on the Y channel of the input images;
    \item \textit{unknown QF} luminance restoration: comparison to test the ability of the models to restore images at intermediate QFs never seen during training; %phase, and to compare with models trained for specific $QF$s;
    \item high and low details density areas restoration: evaluation of the performances of the state-of-the-art methods and the proposed one over specific areas of the images, by dividing the images in patches classified on high-to-low frequency (DCT domain) and high-to-low detail density;
    \item color restoration: evaluation of the color restoration capability of the model on the images converted in RGB space after the elaboration.
\end{enumerate}

%In the following we are going to give a detailed description of the datasets used for training validation and tests, the evaluation metrics used for the comparisons and the training details.

\subsection{Dataset}

The dataset used for training is the \textsc{DIV2K} dataset, a collection of high-quality images (2K resolution), presented during the NTIRE2017 challenge \cite{Agustsson_2017_CVPR_Workshops} for image restoration tasks. This dataset is made of a total amount of 900 images: 800 are used for training while the remaining 100 are used for validation.
The complete dataset contains also 100 images for testing. The groundtruths of this last part have not been released after the challenge, and therefore are not used in this paper.

With the purpose of increase the amount of different texture and pattern to show to the model during training, we have combined the \textsc{DIV2K} dataset with the \textsc{Flickr2K} dataset \cite{timofte2017ntire}, a collection of 2650 high-quality images (same resolution as the \textsc{DIV2K}) collected from Flickr website. 

In order to train the models on different quality factors, for each image in the dataset we have applied 10 different compression levels, corresponding to the quality factors between $\rm{QF}=10$ to $\rm{QF}=100$, with step $10$. 
The images have been compressed in RGB space with the MATLAB standard library function, then the compressed images have been converted later in YCbCr space using the \textsc{Python Scikit-Image} library ($v0.14.0$), during the training phase. The compressed version of the training dataset contains 8000 images. The same operation has been applied to the \textsc{Flickr2K} dataset for a total amount of 34k training images.

The evaluation of our model has been done on the \textsc{LIVE1}\cite{wang2004image}, \textsc{Classic-5}  and \textsc{BSD500} \cite{arbelaez2011contour}, three benchmark datasets widely used for JPEG artifact removal algorithm evaluation. 
For the evaluation of the behaviour of the models with the \textit{unknown compression quality factor} we adopted the \textsc{SDIVL} \cite{corchs2014no-reference}, a dataset proposed for Image Quality Assessment task.

\subsection{Evaluation metrics}

The globally adopted metrics for the evaluation of the quality of images in artifact removal tasks are PSNR, PSNR-B \cite{yim2011quality} (which focus the evaluation on the blocketization in the image) and SSIM \cite{wang2004image}  indexes. For all of these three measures an higher value means better results.
The PSNR and PSNR-B indexes give information about the quality of the images in terms of noise and perceived quality, with PSNR-B taking in consideration also the blocketization artifacts; SSIM index is an indicator of the quality of edges and structures contained in the. For all the three indexes considered an higher value means that the content and the structures in the reconstructed image are more similar to the ones in the target image.

%% Nel trafiletto iniziale?
%\textcolor{red}{Since the state-of-the-art methods operate only on the Y channel of the images, in order to make a fair comparison, the metrics are evaluated on the Y channel recovered by the first network with the corresponding target images, using the MATLAB standard libraries, over five different compression qualities: $10, 20, 40, 60, 80$. We adopted the \textsc{LIVE1}, \textsc{BSD500} and \textsc{Classic-5} datasets, reporting the results from the corresponding publication for each methods, except for ARCNN and MWCNN which provide the source-code, that are then used for the evaluation. Since the training of the proposed methods leads to a single model that can be used for all the quality factors, we used the same model for the evaluation at all the qualities previously mentioned. }

%\textcolor{red}{We also compared the behaviour of the models on unknown quality factors. Since previous models have been trained on specific quality factors, and our model has been instead trained over quality factor from 10 to 100, without the use of images with $QF$s in between, we decided to test the model robustness on ``never seen" artifacts. This evaluation has been done using the \textsc{SDIVL} dataset, where each image has been compressed for all the quality factors between $\rm{QF}=5$ and $\rm{QF}=25$ with step 1. The details are shown in the next section.}

\subsection{Training Details}

All the training phase has been done on a NVIDIA GTX 1070 GPU with $8$ GB of memory using \textsc{PyTorch} framework at version 0.4.1.  
The mini-batch size has been set to $8$ and each input image has been cropped to a patch size of $100\times100$ pixels. During the experiments we tried to train the network with different crop sizes ($32\times32$, $50\times50$, $100\times100$ and $400\times400$), observing how training deeper networks with bigger patch size gives a boost on performances over both PSNR and SSIM indexes. 

We also explored the use of different numbers of RRDBs in the model: we observed how with deeper models, using this specific kind of residual blocks, the results got better and better, increasing the PSNR and SSIM values on the validation set.
The final structure uses five RRDBs for the Y channel restoration model and three RRDBs for the CbCr model, where each convolution has $64$ filters. We found this configuration to be the best one, with respect to the patch size, the amount of RRDBs, the number of filters and the limits due to the memory offered by our board.

We trained the model using Adam optimizer \cite{kingma2014adam} with $\beta_1 = 0.9, \beta_2=0.999$, with learning rate initialized at $2 \times 10^{-4}$ decreased after $200$ epochs of training by a factor of $2$. The training has been performed using the L1 Loss, since allow us to achieve better PSNR results and to make the training more stable.

\section{Experimental~Results}

%\begin{figure}[!t]
%\centering
%\includegraphics[width=\linewidth]{images/qf_20.pdf}
%\caption{PSNR-SSIM comparison of the state-of-the-art-models and our proposed method. For both metrics higher value means better visual results.}
%\label{fig_qf_20}
%\end{figure}
%\begin{figure}[!t]
%\includegraphics[width=\linewidth]{images/qf_40.pdf}
%\caption{PSNR-SSIM comparison of the state-of-the-art-models and our proposed method. For both metrics higher value means better visual results.}
%\label{fig_qf_40}
%\end{figure}

\begin{figure*}[!t]
\centering
\resizebox{\textwidth}{!}{
\begin{tabular}{cc}
\includegraphics[width=\linewidth]{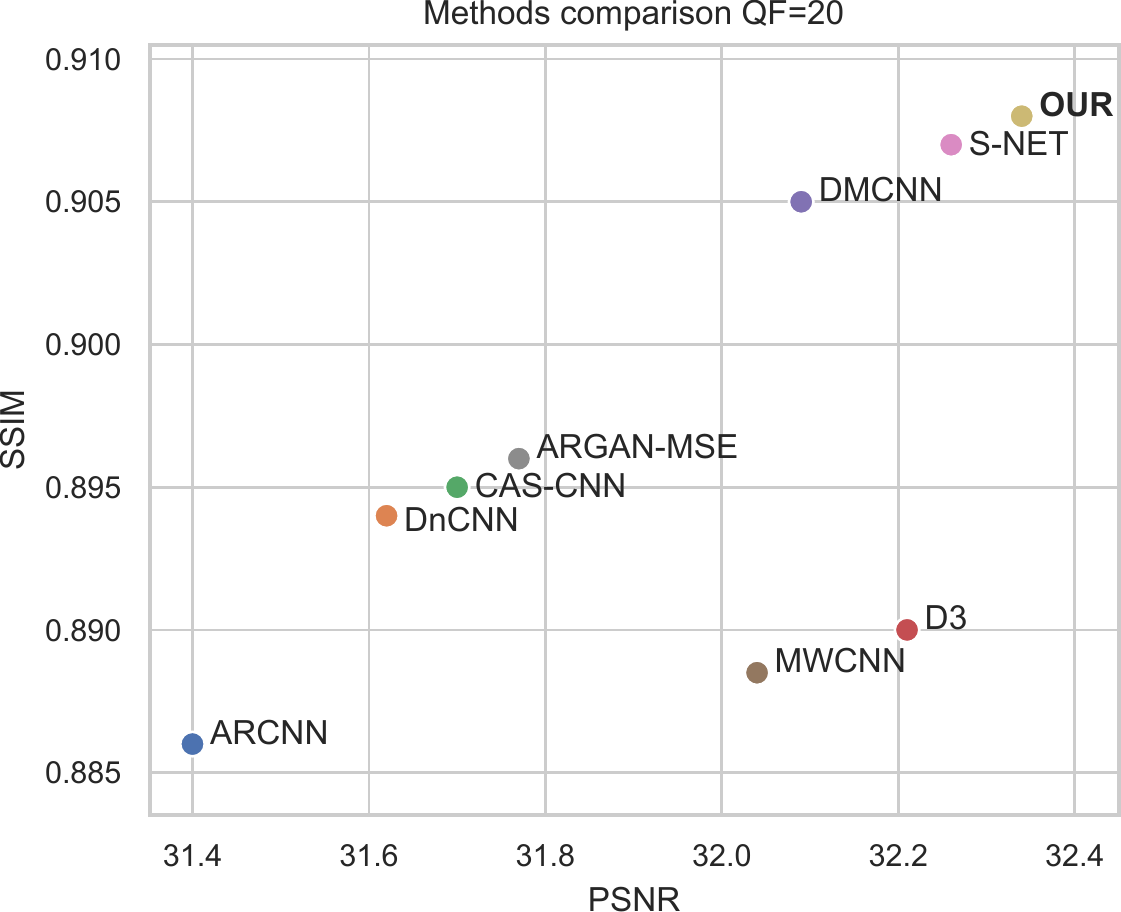} &
\includegraphics[width=\linewidth]{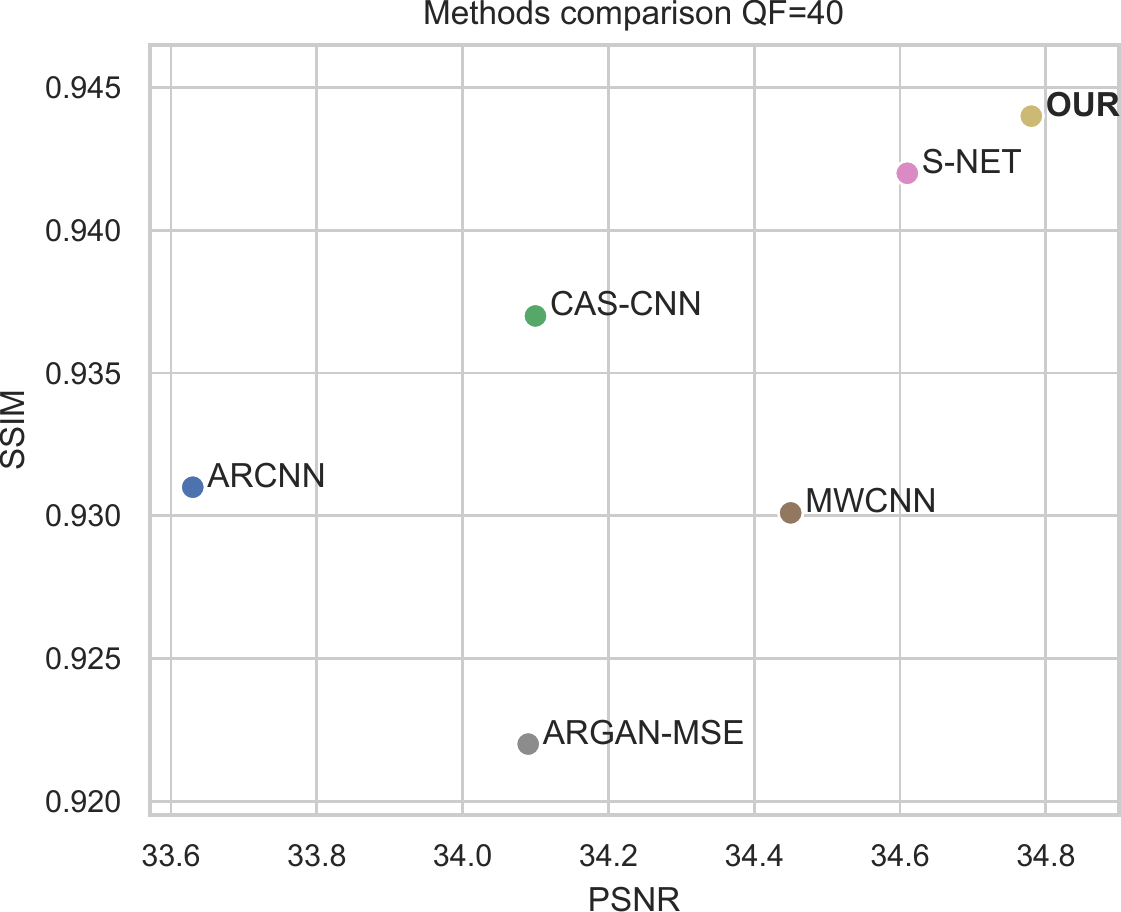} \\
\end{tabular}}
\caption{PSNR-SSIM comparison of the state-of-the-art-models and our proposed method. For both metrics higher value means better visual results.}
\label{fig_qf_20_40}
\end{figure*}

\subsection{Restoration with known compression Quality Factor}

We compared our model with the state-of-the-art models ARCNN\cite{dong2015compression}, CAS-CNN\cite{cavigelli2017cas}, D3\cite{wang2016d3}, and the more recent DMCNN\cite{zhang2018dmcnn}, MWCNN\cite{Liu_2018_CVPR_Workshops}, ARGAN\cite{galteri2017deep} and S-Net\cite{zheng2018s}.
%Since ARCNN and MWCNN are the only ones to provide the open-source code and the pretrained model weights, the evaluation of these models has been done by using the pretrained models over the test sets by ourself; for the other models we used the results reported in the corresponding publications.

Since the state-of-the-art methods operate only on the Y channel of the images, in order to make a fair comparison, the metrics are evaluated on the Y channel recovered by the first network with the corresponding target images, using the MATLAB standard libraries, over five different compression qualities: $10, 20, 40, 60, 80$. %We adopted the \textsc{LIVE1}, \textsc{BSD500} and \textsc{Classic-5} datasets,
For each method, on all the datasets considered, we report the results taken from the corresponding publication, except for ARCNN and MWCNN which provide the source-code, that are then used for the evaluation.
Since the training of the proposed methods leads to a single model that can be used for all the quality factors, we used the same model for the evaluation at all the qualities previously mentioned. All the state-of-the-art methods compared, instead, have a different trained model for each QF considered.

Table \ref{tab_comp_live}, \ref{tab_comp_bsd} and \ref{tab_comp_class} respectively report the comparison on the \textsc{LIVE1}, \textsc{BSD500} and \textsc{Classic-5} datasets for all the three metrics considered.
As can be seen our model outperforms the state of the art on all the metrics. With the proposed model we obtained improvements with respect to the state-of-the-art methods on both general perceptual quality (PSNR/PSNR-B) and structure reconstruction (SSIM).
%\textcolor{red}{The PSNR index gives information about the quality of the images in therms of noise and perceived quality (PSNR-B take in consideration also the blocketization artifacts) while the SSIM index is an indicator of the quality of edges and structures contained in a image with respect to a target one. For both of the indexes an higher value means that the content and the structures in the reconstructed image are more similar to the ones in the target. }
Since each index focuses of different aspects of the restoration quality, each index
%Since the indexes 
alone is not capable to summarize all the aspect of a good reconstruction. Therefore, we also compare the methods in a graph style-view, reported in Figures %\ref{fig_qf_10}, \ref{fig_qf_20} and \ref{fig_qf_40}, 
\ref{fig_qf_10} and \ref{fig_qf_20_40}  
to correlate the two indexes. In order for a method to obtain a more pleasing perceived quality, it is necessary that both the metrics obtain high values. It is easy from this kind of view to see how the proposed method outperforms the current state-of-the-art models even if a single model is used for all the QFs.

\begin{table*}[!t]
\caption{Comparison on test set \textsc{LIVE1}: for the methods in the state of the art a five different models are trained for each QF considered. The proposed method uses the same model for all the QFs.}
\centering
\begin{tabular}{llcccccccccc}
       & Quality   
       & \begin{tabular}[c]{@{}c@{}}ARCNN\\ \cite{dong2015compression}\end{tabular} 
       & \begin{tabular}[c]{@{}c@{}}DnCNN\\ \cite{zhang2017beyond}\end{tabular} 
       & \begin{tabular}[c]{@{}c@{}}CAS-CNN\\ \cite{cavigelli2017cas}\end{tabular}  
       & \begin{tabular}[c]{@{}c@{}}D3\\ \cite{wang2016d3}\end{tabular}  
       & \begin{tabular}[c]{@{}c@{}}DMCNN\\ \cite{zhang2018dmcnn}\end{tabular}  
       & \begin{tabular}[c]{@{}c@{}}MWCNN\\ \cite{Liu_2018_CVPR_Workshops}\end{tabular}
       & \begin{tabular}[c]{@{}c@{}}S-NET\\ \cite{zheng2018s} \end{tabular} 
       & \begin{tabular}[c]{@{}c@{}}ARGAN-MSE\\ \cite{galteri2017deep} \end{tabular}
       & \begin{tabular}[c]{@{}c@{}}ARGAN\\ \cite{galteri2017deep} \end{tabular}
       & OUR   \\\hline
       & 10 & 29.13 & 29.19 & 29.44 & 29.96  & 29.73 & 29.37 & 29.87 & 29.45 & 27.29 & \textbf{29.97} \\
       & 20 & 31.4  & 31.59 & 31.70 & 32.21  & 32.09 & 31.58 & 32.26 & 31.77 & 28.35 & \textbf{32.34} \\
PSNR   & 40 & 33.63 & 33.96 & 34.10 & -      & -     & 34.17 & 34.61 & 34.09 & 28.99 & \textbf{34.78} \\
       & 60 & -     & -     & 35.78 & -      & -     & -     & -     & -     & -     & \textbf{36.47} \\
       & 80 & -     & -     & 38.55 & -      & -     & -     & -     & -     & -     & \textbf{39.31} \\\hline
       & 10 & 28.74 & -     & 29.19 & 29.45  & 29.55 & 28.85 & -     & 29.10 & 26.69 & \textbf{29.60} \\
       & 20 & 30.69 & -     & 30.88 & 31.35  & 31.32 & 30.83 & -     & 31.26 & 28.10 & \textbf{31.76} \\
PSNR-B & 40 & 33.12 & -     & 33.68 & -      & -     & 33.33 & -     & 33.40 & 28.84 & \textbf{33.96} \\
       & 60 & -     & -     & 35.10 & -      & -     & -     & -     & -     & -     & \textbf{35.51} \\
       & 80 & -     & -     & 37.73 & -      & -     & -     & -     & -     & -     & \textbf{38.26} \\\hline
       & 10 & 0.823 & 0.812 & 0.833 & 0.823  & 0.842 & 0.832 & 0.847 & 0.834 & 0.773 & \textbf{0.850} \\
       & 20 & 0.886 & 0.880 & 0.895 & 0.890  & 0.905 & 0.891 & 0.907 & 0.896 & 0.817 & \textbf{0.908} \\
SSIM   & 40 & 0.931 & 0.924 & 0.937 & -      & -     & 0.936 & 0.942 & 0.922 & 0.837 & \textbf{0.944} \\
       & 60 & -     & -     & 0.954 & -      & -     & -     & -     & -     & -     & \textbf{0.960} \\
       & 80 & -     & -     & 0.973 & -      & -     & -     & -     & -     & -     & \textbf{0.976} \\\hline
\end{tabular}
\label{tab_comp_live}
\end{table*}

\begin{table*}[!t]
\caption{Comparison on test set \textsc{BSD500}: for the methods in the state of the art a five different models are trained for each QF considered. The proposed method uses the same model for all the QFs.}
\centering
\begin{tabular}{llcccccccccc}
       & Quality   
       & \begin{tabular}[c]{@{}c@{}}ARCNN\\ \cite{dong2015compression}\end{tabular} 
       & \begin{tabular}[c]{@{}c@{}}DnCNN\\ \cite{zhang2017beyond}\end{tabular} 
       & \begin{tabular}[c]{@{}c@{}}CAS-CNN\\ \cite{cavigelli2017cas}\end{tabular}  
       & \begin{tabular}[c]{@{}c@{}}D3\\ \cite{wang2016d3}\end{tabular}  
       & \begin{tabular}[c]{@{}c@{}}DMCNN\\ \cite{zhang2018dmcnn}\end{tabular}  
       & \begin{tabular}[c]{@{}c@{}}MWCNN\\ \cite{Liu_2018_CVPR_Workshops}\end{tabular}
       & \begin{tabular}[c]{@{}c@{}}S-NET\\ \cite{zheng2018s} \end{tabular} 
       & \begin{tabular}[c]{@{}c@{}}ARGAN-MSE\\ \cite{galteri2017deep} \end{tabular}
       & \begin{tabular}[c]{@{}c@{}}ARGAN\\ \cite{galteri2017deep} \end{tabular}
       & OUR   \\\hline
       & 10 & 29.10 & - & - & - & 29.67 & 29.50 & 29.82 & 29.03 & 27.01 & \textbf{29.92} \\
PSNR   & 20 & 31.25 & - & - & - & 31.98 & 31.34 & 32.15 & 31.20 & 28.07 & \textbf{32.23} \\
       & 40 & 33.55 & - & - & - & -     & 33.23 & 34.45 & 33.30 & 28.61 & \textbf{34.61} \\\hline
       & 10 & 28.75 & - & - & - & -     & 28.60 & -     & 28.61 & 26.30 & \textbf{29.41} \\
PSNR-B & 20 & 30.60 & - & - & - & -     & 29.84 & -     & 30.48 & 27.76 & \textbf{31.39} \\
       & 40 & 32.80 & - & - & - & -     & 31.04 & -     & 32.18 & 28.20 & \textbf{33.34} \\\hline
       & 10 & 0.819 & - & - & - & 0.840 & 0.835 & 0.844 & 0.807 & 0.746 & \textbf{0.847} \\
SSIM   & 20 & 0.885 & - & - & - & 0.904 & 0.889 & 0.905 & 0.876 & 0.794 & \textbf{0.906} \\
       & 40 & 0.929 & - & - & - & -     & 0.928 & 0.941 & 0.921 & 0.815 & \textbf{0.943} \\\hline
\end{tabular}
\label{tab_comp_bsd}
\end{table*}

\begin{table*}[!t]
\caption{Comparison on test set \textsc{Classic-5}: for the methods in the state of the art a five different models are trained for each QF considered. The proposed method uses the same model for all the QFs.}
\centering
\begin{tabular}{llcccccccccc}
       & Quality   
       & \begin{tabular}[c]{@{}c@{}}ARCNN\\ \cite{dong2015compression} \end{tabular} 
       & \begin{tabular}[c]{@{}c@{}}DnCNN\\ \cite{zhang2017beyond} \end{tabular} 
       & \begin{tabular}[c]{@{}c@{}}CAS-CNN\\ \cite{cavigelli2017cas} \end{tabular}  
       & \begin{tabular}[c]{@{}c@{}}D3\\ \cite{wang2016d3} \end{tabular}  
       & \begin{tabular}[c]{@{}c@{}}DMCNN\\ \cite{zhang2018dmcnn} \end{tabular}  
       & \begin{tabular}[c]{@{}c@{}}MWCNN\\ \cite{Liu_2018_CVPR_Workshops} \end{tabular}
       & \begin{tabular}[c]{@{}c@{}}S-NET\\ \cite{zheng2018s} \end{tabular} 
       & \begin{tabular}[c]{@{}c@{}}ARGAN-MSE\\ \cite{galteri2017deep} \end{tabular}
       & \begin{tabular}[c]{@{}c@{}}ARGAN\\ \cite{galteri2017deep} \end{tabular}
       & OUR   \\\hline
       & 10        &  29.04 & 29.4  & -       & -  & -     & \textbf{29.68} & -           & -     & -     & 29.67  \\
PSNR   & 20        &  31.16 & 31.63 & -       & -  & -     & 31.78 & -      & -           & -     & \textbf{31.89} \\
       & 40        &  33.34 & 33.77 & -       & -  & -     & \textbf{34.05} & -           & -     & -     & 34.04  \\\hline
       & 10        &  28.75 & -     & -       & -  & -     & 29.06 & -      & -           & -     & \textbf{29.35} \\
PSNR-B & 20        &  30.6  & -     & -       & -  & -     & 30.95 & -      & -           & -     & \textbf{31.43} \\
       & 40        &  32.8  & -     & -       & -  & -     & 33.20 & -      & -           & -     & \textbf{33.33} \\\hline
       & 10        &  0.811 & 0.803 & -       & -  & -     & 0.828 & -      & -           & -     & \textbf{0.829} \\
SSIM   & 20        &  0.869 & 0.861 & -       & -  & -     & 0.878  & -      & -           & -     & \textbf{0.882} \\
       & 40        &  0.91  & 0.9   & -       & -  & -     & 0.916 & -      & -           & -     & \textbf{0.917} \\\hline
\end{tabular}
\label{tab_comp_class}
\end{table*}

\subsection{Restoration with unknown compression Quality Factor}

Another kind of evaluation has been done about the capability of the models to recover images at compression quality factors never seen during training. In most of the real use-cases, the JPEG compression quality factor previously applied on an image is not know: it is then important that a  model is able to recover the images without this prior information. On the other hand, if we are able at least to estimate the compression quality factor of the input compressed image, following the previous approaches we should train new models for each specific quality factor needed, or use the model trained for the closest QF to the desired one.

%To evaluate how a \textit{general-QF training} impact on the design of a more versatile model for every cases, 
We compare our model with the two state-of-the-art models for which the code i available (i.e. ARCNN and MWCNN) in a specific selection of cases. 
Since  previous  models  have  been  trained  on specific quality factors, and our model has been instead trained over quality factor from 10 to 100 in steps of 10, without the use of images with QFs in between, we decided to test the model robustness on “never seen” artifacts.
%We focus the attention on the quality factors from $5$ to $25$ with step 1, on values never seen by the models during training (this consideration is valid also for our model, trained on qualities from 10 to 100 with step 10). 
In order to perform the evaluation in a coherent way, for the state-of-the-art algorithm we used the pretrained models for the nearest quality factor, for example if the input image has been compressed with $\rm{QF}=17$ we used the models trained for $\rm{QF}=20$. For this evaluation we adopted the \textsc{SDIVL} dataset: for each image of the testset we applied all of the compression factors in the interval $5-25$.
The evaluation is done in the same way it has been done for in the previous secsion, by extracting \textit{Y} channel and measuring PSNR, PSNR-B and SSIM indexes.

\begin{figure*}[!t]
\centering
\hfill
\subfloat[PSNR]{\includegraphics[width=0.33\linewidth]{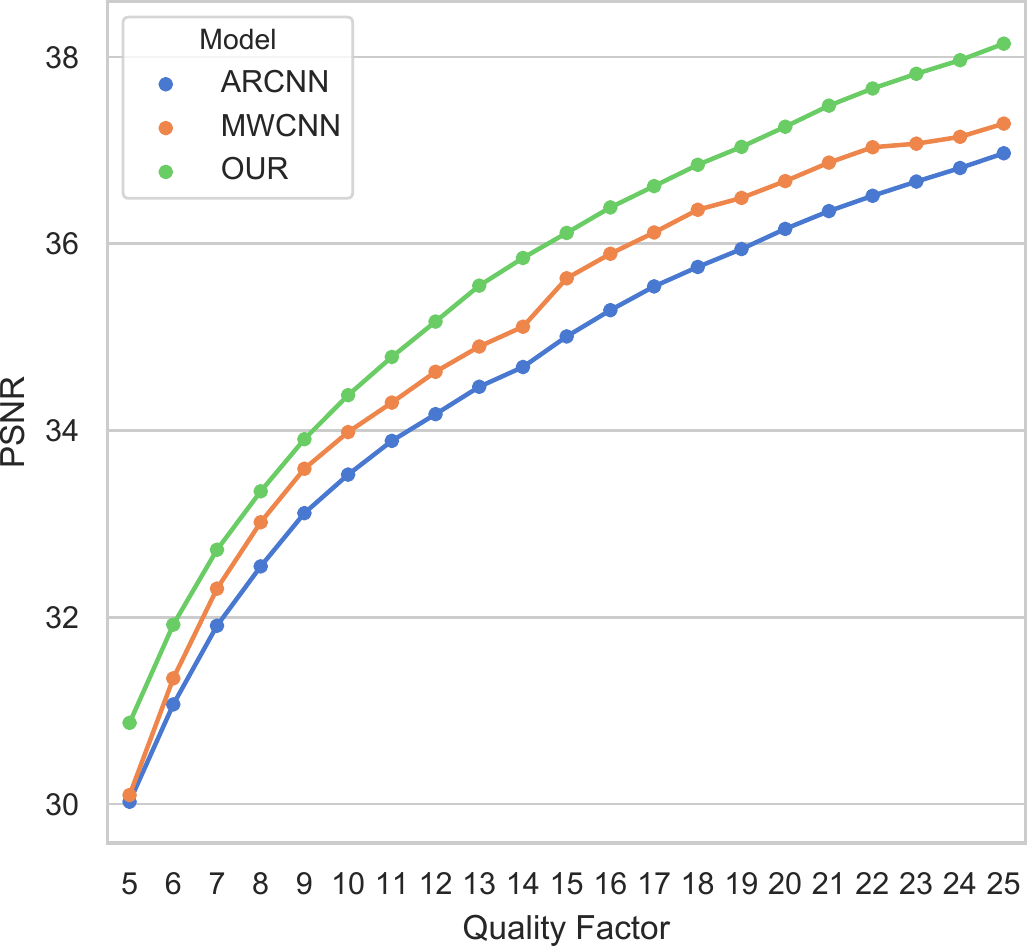}%
\label{fig_psnr_wild}}\hfill
\subfloat[PSNR-B]{\includegraphics[width=0.33\linewidth]{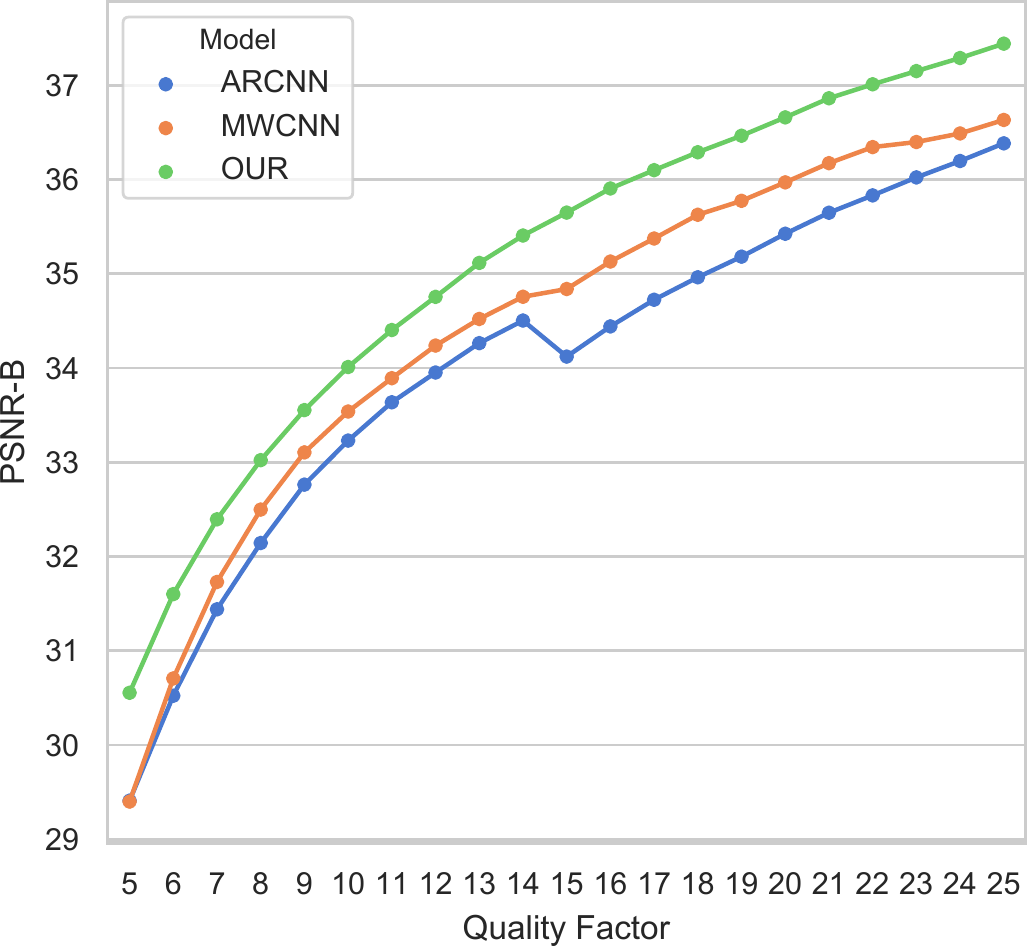}%
\label{fig_psnrb_wild}}\hfill
\subfloat[SSIM]{\includegraphics[width=0.33\linewidth]{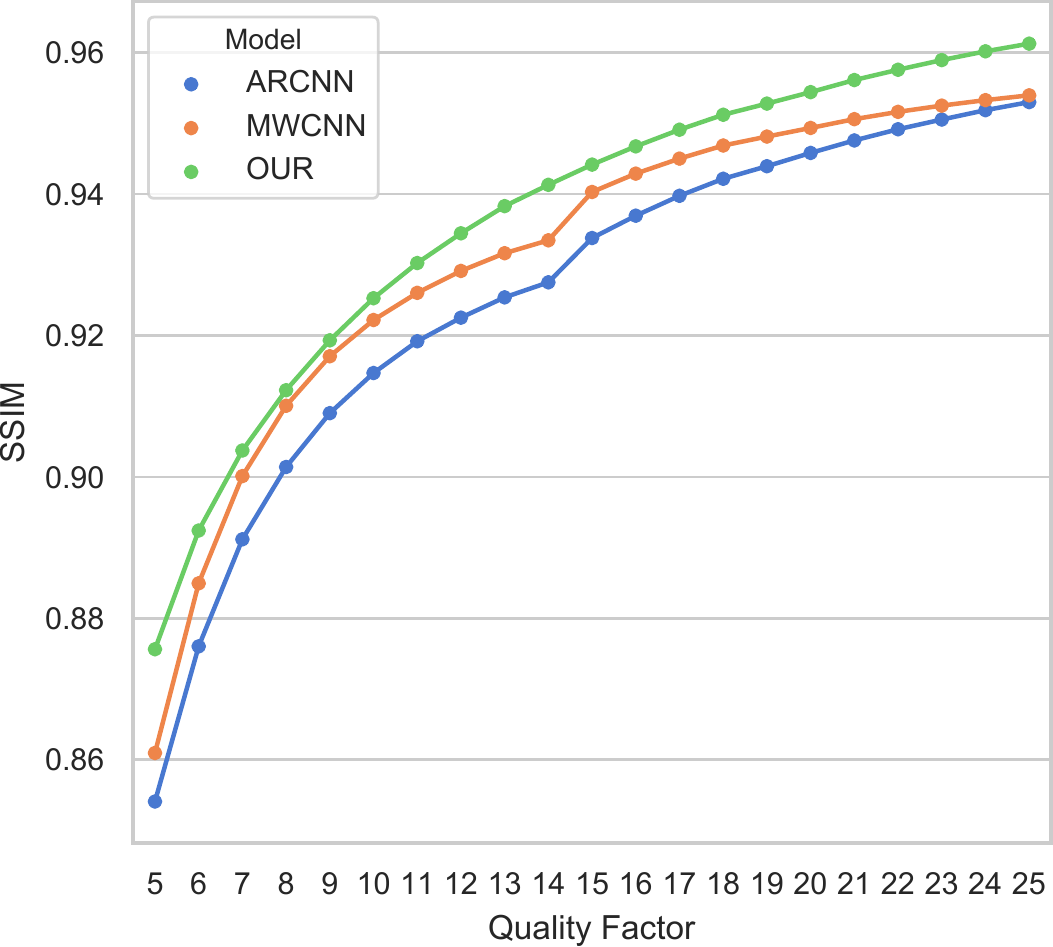}%
\label{fig_ssim_wild}}\hfill

\caption{Comparison on QFs not seen during training. For ARCNN and MWCNN the models trained for QF=10 and QF=20 are tested on QF in the range [5, 25]. The proposed model is trained for QF in the range [10, 100] with steps of 10, and is tested on the same intermediate QFs not seen in training.}
\label{fig_com_inter}
\end{figure*}

In Figure \ref{fig_com_inter} are shown the results of the models on the \textsc{SDIVL} with all the quality factors compression.
As can be seen in those graphs our model shows a more stable behaviour: the model is capable to restore images at different QFs with a more coherent and smooth behaviour in relation to the increase of the QF, in comparison with the other methods. Moreover, the previous state-of-the-art models have difficulties to restore images at quality factors distant from the trained one.
%, as can be seen in the critial points like $QF=14,15$ and $25$. 
It is particularly interesting to see how the other models have difficulties to restore images at higher qualities with respect to the QF used in training, in terms of structures in the images (Figure \ref{fig_ssim_wild}), due to the more complex textures never seen by the models during training phase.

% \begin{figure*}[!t]
% \centering
% \includegraphics[width=\linewidth]{images/sdivl.pdf}
% \caption{results intermediate values.}
% \label{fig_com_inter}
% \end{figure*}

\subsection{High and low frequency areas restoration}
In order to better understand if the proposed method performs better than approaches in the state of the art only on certain image types, we conduct a further experiment: we divide the images from \textsc{LIVE1} testset, compressed at $\textit{QF} = 10$, into $64\times 64$ patches and classify each of them into five categories. The categories are obtained by equally diving the patches into five bins with respect to both frequency and detail density. Patch frequency is computed as the weighted average of the 2D Fourier Transform normalized magnitude. Patch detail density is computed as the 2D average of the result of the Canny edge detection.
The results for the considered evaluation metrics over the five categories of the frequency and detail density are respectively reported in Table \ref{tab_frequency} and \ref{tab_edge}. From the results reported it is possible to notice that the proposed method consistently outperforms the state of the art on all the frequency and detail density categories.

\begin{table*}[!t]
\centering
\caption{Comparison by subdividing the image patches on the basis of the frequency content in five classes from high to low.}
\begin{tabular}{lccccccccc}
                  & \multicolumn{3}{c}{ARCNN} & \multicolumn{3}{c}{MWCNN} & \multicolumn{3}{c}{OUR}                          \\
Frequency         & PSNR   & PSNR-B  & SSIM   & PSNR   & PSNR-B  & SSIM   & PSNR           & PSNR-B         & SSIM           \\\hline
high              & 27.53  & 27.26   & 0.782  & 27.61  & 27.24   & 0.792  & \textbf{28.18} & \textbf{27.88} & \textbf{0.807} \\
medium-high       & 25.00  & 24.66   & 0.685  & 25.24  & 24.67   & 0.700  & \textbf{25.64} & \textbf{25.18} & \textbf{0.729} \\
medium            & 24.61  & 24.27   & 0.734  & 24.50  & 23.82   & 0.740  & \textbf{25.37} & \textbf{24.91} & \textbf{0.773} \\
medium-low        & 25.92  & 25.49   & 0.794  & 25.91  & 25.24   & 0.803  & \textbf{26.73} & \textbf{26.21} & \textbf{0.827} \\
low               & 27.08  & 25.93   & 0.840  & 26.72  & 25.27   & 0.849  & \textbf{27.81} & \textbf{26.52} & \textbf{0.864} \\\hline
\end{tabular}
\label{tab_frequency}
\end{table*}

\begin{table*}[!t]
\centering
\caption{Comparison by subdividing the image patches on the basis of the detail density in five classes from low to high.}
\begin{tabular}{lccccccccc}
                & \multicolumn{3}{c}{ARCNN} & \multicolumn{3}{c}{MWCNN} & \multicolumn{3}{c}{OUR}                          \\
Edges frequency & PSNR   & PSNR-B  & SSIM   & PSNR   & PSNR-B  & SSIM   & PSNR           & PSNR-B         & SSIM           \\\hline
high            & 23.20  & 22.94   & 0.667  & 23.42  & 22.82   & 0.683  & \textbf{23.87} & \textbf{23.45} & \textbf{0.716} \\
medium-high     & 24.69  & 24.39   & 0.721  & 24.91  & 24.31   & 0.735  & \textbf{25.42} & \textbf{25.02} & \textbf{0.763} \\
medium          & 25.68  & 25.22   & 0.758  & 25.94  & 25.26   & 0.772  & \textbf{26.41} & \textbf{25.86} & \textbf{0.794} \\
medium-low      & 26.83  & 26.12   & 0.805  & 27.01  & 25.95   & 0.817  & \textbf{27.47} & \textbf{26.61} & \textbf{0.832} \\
low             & 29.17  & 28.22   & 0.884  & 27.45  & 26.28   & 0.888  & \textbf{29.97} & \textbf{28.99} & \textbf{0.897} \\\hline
\end{tabular}
\label{tab_edge}
\end{table*}

\subsection{Color Restoration}

The final evaluation is focused on the color restoration capability of the models. The comparison, in the same way as done in the previous evaluations, has been done among the ARCNN\cite{dong2015compression}, MWCNN\cite{Liu_2018_CVPR_Workshops} and our proposed model.

We restored the images from the \textsc{LIVE1} testset with the lowest quality factors $\rm{QF}=10,20,40$. For this specific evaluation we restored both luma and chroma components. In the case of ARCNN and MWCNN methods, we adopted the same model for all of the three channels (\textit{Y}, \textit{Cb} and \textit{Cr} channels), while our method uses the two different networks to first restore the luminance channel then the chrominance channels.

\begin{figure*}[!t]
\centering
\resizebox{1.00\textwidth}{!}{
\begin{tabular}{ccccc}
\includegraphics[width=0.18\linewidth]{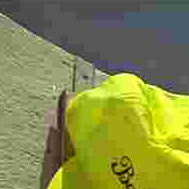} &
\includegraphics[width=0.18\linewidth]{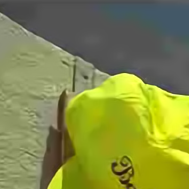} & 
\includegraphics[width=0.18\linewidth]{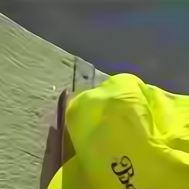} &
\includegraphics[width=0.18\linewidth]{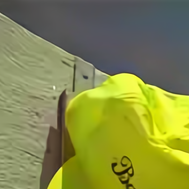} & 
\includegraphics[width=0.18\linewidth]{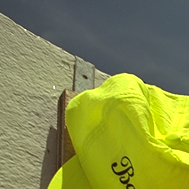} \\

\includegraphics[width=0.18\linewidth]{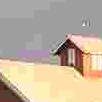} &
\includegraphics[width=0.18\linewidth]{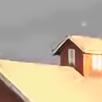}& 
\includegraphics[width=0.18\linewidth]{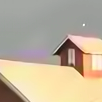} &
\includegraphics[width=0.18\linewidth]{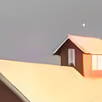} &
\includegraphics[width=0.18\linewidth]{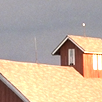} \\

\includegraphics[width=0.18\linewidth]{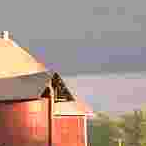} &
\includegraphics[width=0.18\linewidth]{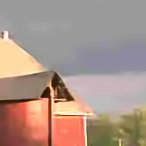} & 
\includegraphics[width=0.18\linewidth]{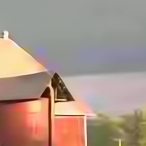} &
\includegraphics[width=0.18\linewidth]{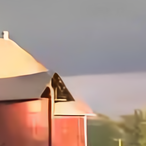} & 
\includegraphics[width=0.18\linewidth]{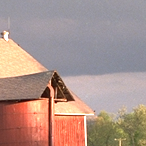} \\

% \subfloat[Input]{\includegraphics[width=0.18\linewidth]{images/comp_color/lighthouse/c1_low.png}} &
% \subfloat[ARCNN]{\includegraphics[width=0.18\linewidth]{images/comp_color/lighthouse/c1_ar.png}\label{fig_color_ar}} & 
% \subfloat[MWCNN]{\includegraphics[width=0.18\linewidth]{images/comp_color/lighthouse/c1_mw.png}\label{fig_color_mw}} &
% \subfloat[OUR]{\includegraphics[width=0.18\linewidth]{images/comp_color/lighthouse/c1_our.png}} &
% \subfloat[Target]{\includegraphics[width=0.18\linewidth]{images/comp_color/lighthouse/c1_or.png}} \\
\includegraphics[width=0.18\linewidth]{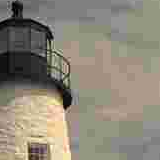} &
\includegraphics[width=0.18\linewidth]{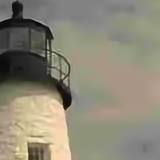} & 
\includegraphics[width=0.18\linewidth]{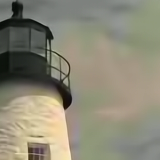} &
\includegraphics[width=0.18\linewidth]{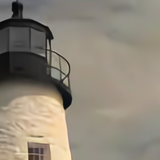} & 
\includegraphics[width=0.18\linewidth]{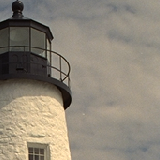} \\
\footnotesize{(a) Input} & \footnotesize{(b) ARCNN} & \footnotesize{(c) MWCNN} & \footnotesize{(d) OUR} & \footnotesize{(e) Target} \\

\end{tabular}
}
\caption{Visual comparison of the full color JPEG restoration.}
\label{fig_color_differences}
\end{figure*}

For this comparison we used the PSNR, PSNR-B and SSIM indexes over the restored images in RGB space, instead of only evaluating the luminance information, using the MATLAB standard library: numerical results can be seen in Table \ref{tab_chroma_comp} and visual results are summarized in some patches from the images of \textsc{LIVE1} in Figure \ref{fig_color_differences}. As can be seen the proposed model obtains better results than the other methods in terms of PSNR, PSNR-B and SSIM index, and is also evident the difference on the final images. The blocketization and the color aberration coming from the compression are blurred and mainteined in the other models, while are cleaned by our model which reshapes the color information with respect to the structures in the images. The results are much more pleasing and realistic than the other methods ones.

\begin{table}[!t]
\centering
\caption{Comparison of the evaluation metrics computer on full-color restorated images}
\begin{tabular}{llccc}
\multicolumn{5}{c}{LIVE1}                           \\
       & Qualities & ARCNN & MWCNN & OUR            \\\hline
       & 10        & 28.97 & 29.84 & \textbf{29.98} \\
PSNR   & 20        & 31.31 & 32.00 & \textbf{32.35} \\
%       & 30        & 32.69 & 33.58 & \textbf{33.78} \\
       & 40        & 33.64 & 34.58 & \textbf{34.80}  \\\hline
       & 10        & 28.69 & 29.40 & \textbf{29.61} \\
PSNR-B & 20        & 30.78 & 31.28 & \textbf{31.77} \\
%       & 30        & 32.15 & 32.91 & \textbf{33.05} \\
       & 40        & 33.13 & 33.78 & \textbf{33.98} \\\hline
       & 10        & 0.822 & 0.846 & \textbf{0.851} \\
SSIM   & 20        & 0.888 & 0.900 & \textbf{0.909} \\
%       & 30        & 0.917 & 0.929 & \textbf{0.932} \\
       & 40        & 0.931 & 0.941 & \textbf{0.944} \\\hline
%                 & 10        & 5.343 & 6.325 & \textbf{5.070} \\
% $\Delta E_{76}$ & 20        & \textbf{3.587} & 4.614 & 3.594 \\
%                 & 40        & \textbf{2.627} & 3.671 & 2.815 \\\hline
       
\end{tabular}
\label{tab_chroma_comp}
\end{table}

\section{Conclusion}
In this paper we proposed a deep residual autoencoder exploiting Residual-in-Residual Dense Blocks (RRDB) to remove artifacts in JPEG compressed images, that is independent from the QF used. 
%The proposed approach exploits both the learning capacity of deep residual networks and the prior knowledge of the JPEG compression pipeline. 
The proposed model operates in the YCbCr color space and performs a two-phase restoration of JPEG artifacts: in the former phase, a first autoencoder exploiting 2D convolutions is used to restore the luma channel; in the latter phase, a  second autoencoder, by stacking along the channel dimension the results of the first autoencoder and the original chroma channels, employs 3D convolutions to exploit the restored luma channel as a guide, and restores the chroma channels.

The main contributions of this paper are: i)  the design of a method for the restoration of JPEG compression artifact that is independent from the QF used; ii) the design of a model trainable end-to-end that fully exploits knowledge about JPEG compression pipeline; iii) a thorough comparison with the state of the art on three standard datasets at fixed QFs; iv) an analysis of robustness of restoration results at QFs not used for training.

Extensive experimental results on three widely used benchmark datasets (i.e. \textsc{LIVE1}, \textsc{BDS500}, and \textsc{CLASSIC-5}) show that our model is able to outperform the state of the art with respect to all the evaluation metrics considered (i.e. PSNR, PSNR-B, and SSIM). This results is remarkable since the approaches in the state of the art use a different set of weights for each compression quality, while the proposed model uses the same weights for all of them, making it applicable to images in the wild where the QF used for compression is unkwnown. Furthermore, the proposed model shows a greater robustness than state-of-the-art methods when applied to compression qualities not seen during training.
Since preliminary experiments with the same architecture proposed showed good results for the restoration of other artifacts (i.e. noise removal, in the CVPRW NTIRE2019 challenge), as future work we plan to investigate its extension to other single and multiple distortions \cite{corchs2017multidistortion}.

% or

% \appendices
% \section{}
% \section*{Acknowledgment}

\ifCLASSOPTIONcaptionsoff
  \newpage
\fi


% \newpage

\bibliographystyle{IEEEtran}
\bibliography{./biblio.bib}


%\end{IEEEbiography}


%\newpage



%\vfill


\end{document}